\documentclass[10pt,twocolumn,letterpaper]{article}
\usepackage{iccv}
\usepackage{times}
\usepackage{epsfig}
\usepackage{graphicx}
\usepackage{amsmath}
\usepackage{amssymb}
\usepackage{xcolor}

\usepackage{times}
\usepackage{epsfig}
\usepackage{amsmath}
\usepackage{amssymb}
\usepackage{textcomp}
\usepackage{multirow}
\usepackage{adjustbox}

\usepackage{multirow}
\usepackage{colortbl}
\usepackage{hhline}
\usepackage{subcaption}
\usepackage{enumitem}
\usepackage{verbatim}
\usepackage{subcaption}

\usepackage{gensymb}
\usepackage[pagebackref=true,breaklinks=true,letterpaper=true,colorlinks,bookmarks=false]{hyperref}

\iccvfinalcopy % *** Uncomment this line for the final submission

\def\iccvPaperID{35} % *** Enter the ICCV Paper ID here

\ificcvfinal\fi

\begin{document}

%%%%%%%%% TITLE
\title{
FuseMODNet: Real-Time Camera and LiDAR based \\ 
    Moving Object Detection for robust low-light Autonomous Driving 
}

\author{Hazem Rashed$^1$, Mohamed Ramzy$^2$, Victor Vaquero$^3$, Ahmad El Sallab$^1$, \\ Ganesh Sistu$^4$  and Senthil Yogamani$^4$\\
{\normalsize $^1$Valeo R\&D, Egypt\hspace{0.3cm}$^2$ Cairo University \hspace{0.3cm}} \\ 
{\normalsize $^3$Institut de Rob\`otica i Inform\`atica Industrial, (CSIC-UPC), Spain \hspace{0.3cm} $^4$Valeo Vision Systems, Ireland}\\
{\tt\small firstname.lastname@valeo.com, mohamed.ibrahim98@eng-st.cu.edu.eg, vvaquero@iri.upc.edu}
%\thanks{Institut de Rob\`otica i Inform\`atica Industrial, CSIC-UPC}
}

% \author{Michal U\v{r}i\v{c}\'{a}\v{r}$^{1}$ and Jan Uli\v{c}n\'{y}$^{3}$ and Pavel K\v{r}\'{i}\v{z}ek$^{1}$, Ganesh Sistu$^{2}$ and David Hurych$^{4}$ and Senthil Yogamani$^{2}$ \\ % <-this % stops a space 
% %\thanks{*This work was not supported by any organization}% <-this % stops a space
% $^{1}$Valeo R\&D Prague, Czech Republic $^{2}$Valeo Visions Systems, Ireland \\
%          {\tt \small \{michal.uricar, pavel.krizek, ganesh.sistu, senthil.yogamani\}@valeo.com}
% }
% \thanks{$^{4}$Valeo ai}
% \thanks{$^{3}$Valeo Bietigheim, Germany}
% \thanks{$^{2}$Valeo Vision Systems, Ireland
%         {\tt\small ganesh.sistu@valeo.com}}%
% \thanks{$^{1}$Valeo R\&D Prague, Czech Republic
%         {\tt\small michal.uricar@valeo.com}}%

% Hazem Rashed 
% Antonin Vobecky

% SPLIT: Michal <- {Abstract, Intro, Soiling_task, Results}
%        Gansesh <- {Architecture, Results}
%        Pavel <- {Architecture, Results}
%        Senthil <- {Abstact, Intro, Soiling_task, Results}
%        Jan
%        David 

\maketitle
\ificcvfinal\thispagestyle{empty}\fi

%%%%%%%%%%%%%%%%%%%%%%%%%%%%%%%%%%%%%%%%%%%%%%%%%%%%%%%%%%%%%%%%%%%
\begin{abstract}
Moving object detection is a critical task for autonomous vehicles. As dynamic objects represent higher collision risk than static ones, our own ego-trajectories have to be planned attending to the future states of the moving elements of the scene. Motion can be perceived using temporal information such as optical flow. Conventional optical flow computation is based on camera sensors only, which makes it prone to failure in conditions with low illumination. On the other hand, LiDAR sensors are independent of illumination, as they measure the time-of-flight of their own emitted lasers. 
In this work, we propose a robust and real-time CNN architecture for Moving Object Detection (MOD) under low-light conditions by capturing motion information from both camera and LiDAR sensors.
We demonstrate the impact of our algorithm on KITTI dataset where we simulate a low-light environment creating a novel dataset ``Dark-KITTI". We obtain a 10.1\% relative improvement on Dark-KITTI, and a 4.25\% improvement on standard KITTI relative to our baselines. The proposed algorithm runs at 18 fps on a standard desktop GPU using $256\times1224$ resolution images.
\end{abstract}

\section{Introduction} \label{sec:intro}
\begin{figure}[ht!]
\centering
    \includegraphics[width=.48\textwidth]{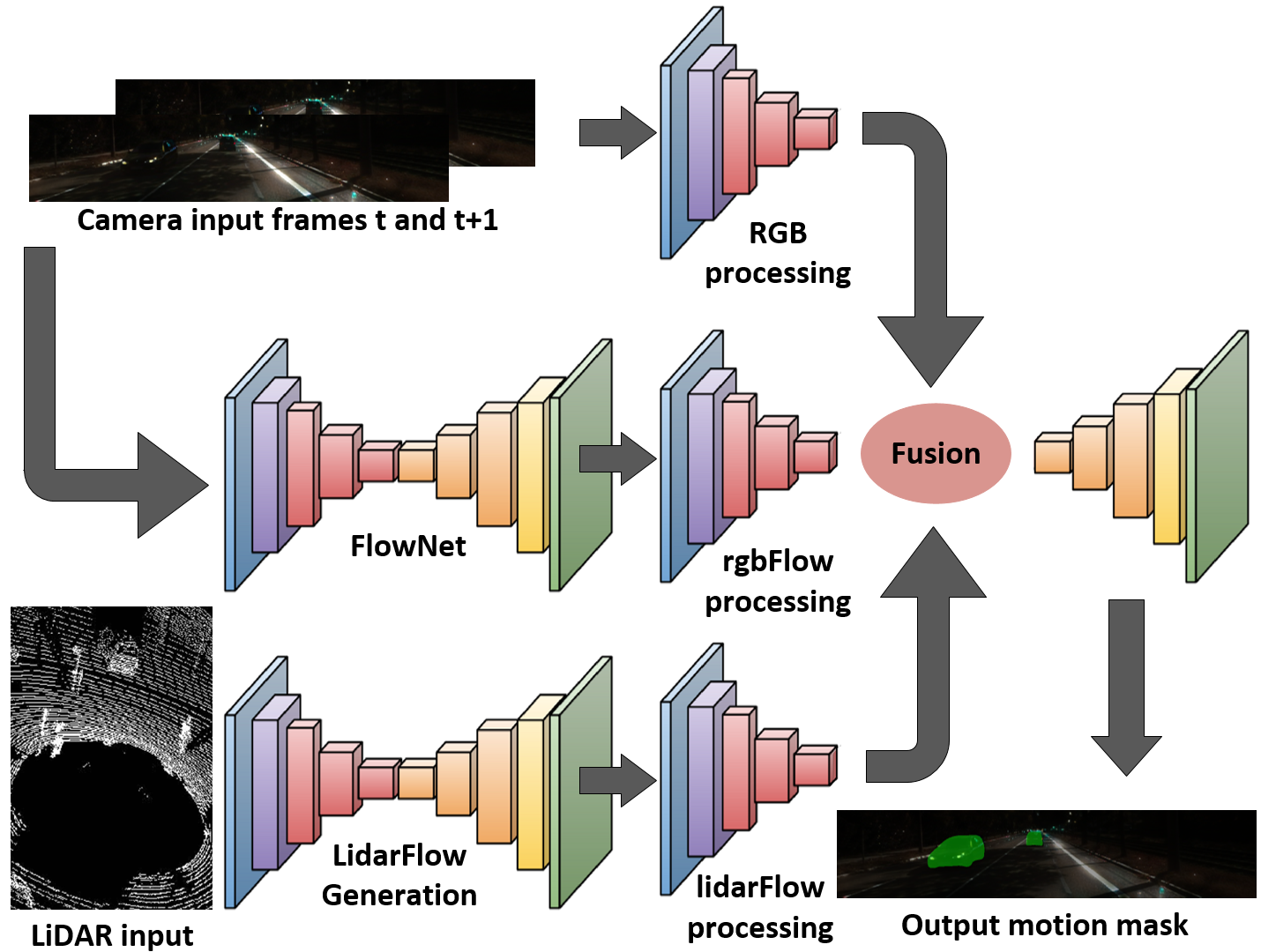}
    \caption{\textcolor{black}{Proposed Network Architecture}
    }
    \vspace{-0.4cm}
\end{figure}%

Autonomous Driving (AD) scenarios are considered very complex environments as they are highly dynamic containing multiple object classes that move at different speeds in diverse directions \cite{horgan2015vision, heimberger2017computer}. For an autonomous vehicle, it is critical to fully understand the motion model of each of the surrounding elements as well as to be able to plan the ego-trajectories based on the future states of these objects, therefore avoiding collision risks. There are two types of motion in a typical autonomous driving scene, i.e.   
%The first one is motion of the surrounding obstacles and the second is motion of the ego-vehicle. Due to the movement of the camera reference itself, it is difficult to successfully classify the surrounding objects as moving or static, because even static objects will be perceived as moving.
the one of the surrounding obstacles and the motion of the ego-vehicle. Due to the movement of the camera reference itself, it is challenging to successfully classify the surrounding objects as moving or static, because even static objects will be perceived as moving.
Motion segmentation implies two tasks that are performed jointly. The first one focuses on object selection, in which objects of specific interesting classes are highlighted such as pedestrians or vehicles. The second one focuses on motion classification, in which a binary classifier predicts whether the observed object is dynamic or static.

Modern vehicles are equipped with various sensors to be able to fully perceive the surrounding environment, each one having its advantages and disadvantages. For instance, ultrasonic sensors provide good performance of depth measurement for close obstacles but they lack semantic information and perform poorly for far objects. 
%Camera sensors instead, provide rich color information from which scene semantics can be extracted however, they lack depth measurement and rely on scene illumination where performance of any camera-based environment perception tasks is usually degraded in bad illumination conditions such as night environment.
Camera sensors instead, provide rich color information from which scene semantics can be extracted however, they lack of depth information and rely on scene illumination, being the performance of any camera-based perception tasks highly degraded in bad illumination conditions such as at night scenes.
On the other hand, LiDAR sensors provide accurate depth and geometric information of the environment, although they generate big and sparse point clouds that may suppose a computational bottleneck. Nevertheless, unlike camera sensors, LiDARs rely on the Time of Flight (ToF) concept and therefore they can perform much better under low illumination or light changing conditions.  

Data fusion has been proven to provide improved performance in various tasks such as \cite{rashed2019motion,siam2017modnet,jain2017fusionseg}.
% There are several types of fusion in ConvNets. Early-Fusion refers to data-fusion where different types of data are fused together before feature extraction. Mid-Fusion refers to feature-fusion where each input data is processed separately and features are extracted from each input then fusion is performed on the feature-level. Late-Fusion refers to fusion on the decision level where the algorithm takes the final decision based on each input separately such as classification, and the output is fused in the end. 
%In this work, we focus on fusion between Camera and LiDAR sensors for the purpose of moving objects detection. Our proposed network attempts to capture semantic information from camera sensor, and motion information from both camera and LiDAR sensors. 
In this work, we focus on fusing Camera and LiDAR information for the purpose of moving objects detection. Our proposed architecture attempts to capture rich motion information from both camera and LiDAR sensors which is combined with scene semantics from the camera images.
To summarize, the contributions of this work include:

\begin{itemize} [nosep]
    \item We extend the publicly available KittiMoSeg~\cite{siam2017modnet} dataset almost x10 times, expanding from 1300 frames only to a new amount of 12919 images. The dataset is available at \url{https://sites.google.com/view/fusemodnet} 
    \item We create the new Dark-KITTI dataset to simulate low illumination autonomous driving environments.
    \item We propose a novel CNN architecture for MOD fusing both RGB and LiDAR information. Our implementation performs on real-time, and therefore is suitable for time-critical applications such as autonomous driving.
    \item We analyze different fusion methodologies for maximum performance as well as study motion representations for both RGB frames and LiDAR points clouds. \\
\end{itemize}

The rest of the paper is organized as follows: a review of
the related work is presented in Section \ref{sec:related}. Our methodology including the dataset preparation and the used network architectures is detailed in Section \ref{sec:method}. Experimental setup and final results are illustrated in Section \ref{sec:exps}. Finally, Section \ref{sec:conclusions} concludes the paper.

\section{Related Work} \label{sec:related}

\textbf{Motion Segmentation using Camera sensor:}
Classical approaches have been proposed for moving objects detection based on geometrical understanding of the scene such as \cite{menze2015object} which was used to estimate objects motion masks. Wehrwein et al. \cite{wehrwein2017video} introduced assumptions about the camera motion model to model the background motion in terms of homography. This approach cannot be used in autonomous driving application due to the errors arising from the limited assumptions such as camera translations. Classical methods provide poor performance compared to deep learning methods in addition to high complexity due to complicated pipelines used. For instance, Menze et al. \cite{menze2015object} running time is 50 minutes per frame which makes it impossible for usage in a real-time application such as the autonomous driving.
Deep learning algorithms are becoming successful beyond object detection \cite{siam2017deep} for applications like visual SLAM \cite{milz2018visual}, depth estimation \cite{kumar2018monocular}, soiling detection \cite{uricar2019soilingnet} but it is still relatively less explored for MOD task.

Jain et. al.\cite{jain2017fusionseg} proposed a method to exploit optical flow for generic foreground segmentation. This work is designed for generic object segmentation and does not focus on classifications of objects as Moving or Static. Drayer et. al.\cite{drayer2016object} proposed a video segmentation algorithm that is based on R-CNN detection. The approach is not practical as well for autonomous driving application due to its complexity where it runs on image in 8 seconds. Siam et al. \cite{siam2017modnet,siam2018real} explored motion segmentation using deep network architectures, however these networks rely only on camera RGB images which is prone to failure in low illumination conditions. FisheyeMODNet \cite{yahiaoui2019fisheyemodnet} extends MODNet for fisheye camera images using WoodScape dataset \cite{yogamani2019woodscape}.

\textbf{Motion Segmentation using LiDAR sensor:}
Most of LiDAR-based methods that have been used for motion segmentation problem were based on clustering methods such as \cite {dewan2016motion} which predicts the points motion by methods such as RANSAC, and then clustering takes place for object-level perception. Vaquero et al. \cite {vaquero2017deconvolutional} initially clustered vehicles points and then performed motion segmentation on the objects after matching the objects through sequential frames. Deep Learning has been utilized in various methods for object detection on point clouds. In \cite {li20173d} 3D convolution is used over the point cloud to obtain the vehicles bounding boxes. Other methods project the 3D points on 2D images to make use of 2D convolutions on the image 2D space \cite {li2016vehicle}. None of these methods are able to segment moving objects from static ones. Recent work \cite{dewan2017deep} learns movable and non-movable objects from two input lidar scans. This method uses implicit learning for motion information through two sequential lidar scans and does not utilize the color information from camera sensor, which motivates our work towards fusion of both camera and LiDAR sensors.

\textbf{Fusion:}
Fusion has been explored through classical and deep learning methods, and it has proven to be very important for many tasks. The most common way for multimodal fusion using classical approaches is Kalman filter \cite{kalman1960new} and its variants. CNNs have been exploited as well for multimodal fusion where they generally provide improved performance over Kalman filters at the cost of complexity. Deep fusion has been explored for the task of semantic segmentation \cite{rashed2019motion,Rashed2019OpticalFA,hazirbas2016fusenet} using fusion between RGB images and optical flow and depth. Several methods have been visited to fuse camera and LiDAR sensors for various tasks such as \cite{1906.00208} which implemented an algorithm for 3D semantic segmentation. Pedestrian detection has been improved significantly using fusion between RGB images and infrared maps \cite{konig2017fully,li2019illumination,wagner2016multispectral}. Modern vehicles are usually equipped with various sensors to perceive the environment which we propose to leverage using a deep fusion network.

% We make use of multimodal fusion to maximize the benefit of each sensor for the task of MOD where we attempt to address the limitation of camera sensor by capturing motion signal via illumination independent LiDAR.

\section{Methodology} \label{sec:method}
In this section we discuss dataset preparation, and the proposed architecture for our experiments.

\subsection{Dataset Preparation} \label{subsec:baseline}
Our proposed method fuses color images with motion signals obtained from different sensors to generate motion masks as output. In this section we describe the inputs preparation and outputs of our architecture. \\

\noindent\textbf{Annotations Generation:}
In order to train our deep model for maximum generalization on the motion segmentation task, we need motion masks annotations from a large driving dataset. There is huge limitation in publicly available datasets regarding moving objects detection. Siam et al. \cite{siam2017modnet} provides 1300 images only with weak annotation for MOD task. Valada et al. \cite{Valada_2017_IROS} provides 255 annotated frames only on KITTI dataset, and 3475 annotated frames on Cityscapes \cite{Cordts2016Cityscapes} dataset. Cityscapes does not have LiDAR point clouds, and therefore will not be helpful for our low-light purposes. 
Behley et al. \cite{behley2019iccv} provides MOD annotations for 3D point clouds only, but not for dense pixels. 
We therefore build our own Motion Object Detection dataset. For that, we adopt the method in \cite{siam2017modnet} to generate motion masks from KITTI in order to extend the KittiMoSeg dataset. Initially, we project the existing 3D bounding boxes from 3D LiDAR frame to 2D pixel coordinate system, as use the given tracking information to compute velocity vectors for each of the surrounding objects in 3D space. In addition, we use GPS readings to compute the ego-vehicle velocity vector for the camera sensor where the difference between both velocities is calculated and compared to a threshold for classifying the objects as moving or static. 
Finally, MaskRCNN \cite{He2017MaskR} segmentation masks are used for refining the obtained output masks. We applied this approach on KITTI-raw frames which have corresponding LiDAR points clouds and tracklets information, obtaining a dataset with a total number of 12919 frames which we split into 80\% for training and 20\% for testing. \\
 
\begin{figure}[t!]
\centering
    \includegraphics[width=.48\textwidth]{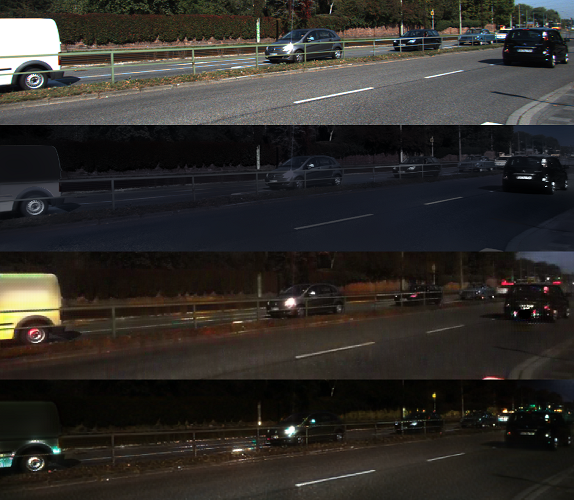}
    \caption{\textcolor{black}{An example of our different night generation methods, \textbf{Top to Bottom:} Input KITTI Image, Neural Style Transfer \cite{li2018closed}, CycleGAN\cite{Zhu-ICCV-2017}, UNIT\cite{liu2017unsupervised}}}
    \vspace{-0.4cm}
\label{nightImages}
\end{figure}

\noindent\textbf{Color Signal:}
Our objective is to develop a complete system for moving object detection to work robustly under any illumination condition. 
%For that purpose we need to evaluate our algorithm on conventional AD scenes in addition to other scenes with low illumination because camera-only based systems usually fail in low-illumination conditions.
For that purpose we require to evaluate our algorithm, in addition to conventional AD scenes, into other more challenging low illumination environments where camera-only based systems would fail due to the lack of textured information.
As far as we know, there exists no dataset providing low-illumination or night scenes in addition to the information needed to generate our MOD annotation. For that reason, we make use of the Image-to-Image translation technique of~\cite{liu2017unsupervised} to generate dark images from the KITTI dataset that mimic night AD scenes. To be able to generate dark realistic frames, we trained UNIT \cite{liu2017unsupervised} network using 2000 KITTI\cite{Geiger2012CVPR} images and 2000 night images from \cite{yu2018bdd100k}. Figure \ref{nightImages} shows a sample of our newly generated dataset which we call Dark-KITTI. It comprises of 12919 night images corresponding to KITTI-raw frames. Other approaches such as \cite{CycleGAN2017,li2018closed} have been attempted to simulate Dark-KITTI images, however we found \cite{liu2017unsupervised} to be more realistic as illustrated in the final row of Figure \ref{nightImages}. \\

\noindent\textbf{Motion Signal:} \label{subsec:motion}
The key input for moving object detection that we give to our system is the motion information obtained from the scene. In order to build an illumination-independent system, we intend to perceive motion from both camera and LiDAR sensors. 
Motion can be either implicitly learned from temporally sequential frames, or provided explicitly to the system through an input motion map, as for example optical flow maps. 
%For RGB images, We make use of FlowNet\cite{Ilg2016FlowNet2E} algorithm for that purpose. 
For obtaining motion from LiDAR information, we leverage a recent approach \cite{vaquero2018hallucinating} that learns to model optical flow maps from LiDAR point clouds. Using this approach, we have the advantage of understanding motion of the surrounding scene even in darkness because LiDAR is illumination independent. 
In addition to these optical flow maps from LiDAR which we term as ``lidarFlow", we generate image-based optical flow using the FlowNet~\cite{Ilg2016FlowNet2E} algorithm over RGB images which we term as ``rgbFlow". 
There is a significant degradation on rgbFlow when it is generated from the Dark-KITTI dataset compared to standard KITTI frames, which is expected given that rgbFlow is illumination-dependent. 

\begin{figure}[t!]
\centering
    \includegraphics[width=.48\textwidth]{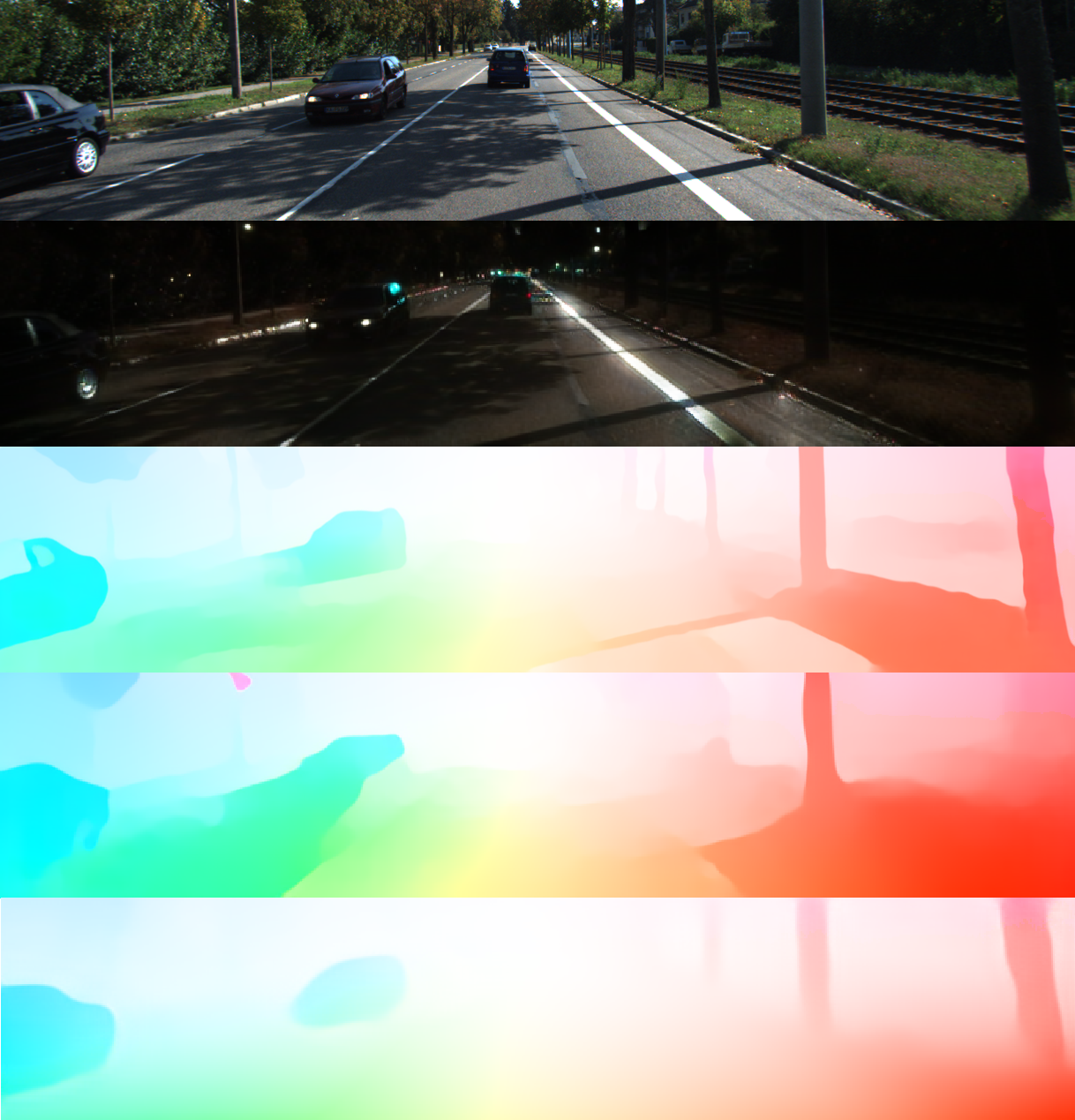}
    \caption{\textcolor{black}{Sample from our Dark-KITTI dataset and the corresponding optical flow images. \textbf{Top to Bottom:} KITTI image; Dark-KITTI image; rgbFlow from KITTI image using FlowNet\cite{Ilg2016FlowNet2E}; degraded rgbFlow on Dark-KITTI image; lidarFlow\cite{vaquero2018hallucinating} obtained just using LiDAR information.}}
    \vspace{-0.4cm}
\label{flowImages}
\end{figure}

In our experiments, we prove that both lidarFlow and rgbFlow are complementary to each other and that the inclusion of LiDAR-based motion signals significantly improve MOD results. In order to align our images with the output from \cite{vaquero2018hallucinating}, we crop the upper part of the dataset frames to be 256x1224 which has no impact on MOD because the moving objects are in the lower part of the image. 
Figure \ref{flowImages} shows a sample of our generated Dark-KITTI dataset along with the corresponding optical flow maps generated from \cite{Ilg2016FlowNet2E,vaquero2018hallucinating}. It can be observed that RgbFlow using high-illumination images during day provide high intensity motion vectors.  However, there exists some distortions such as the ones due to shadow on the ground as illustrated on the 3rd row in Figure \ref{flowImages}, where shadow pixels are perceived as moving pixels and also combined with the moving pixels from the cars. 
For low-illumination rgbFlow in the fourth row, it can be appreciated that it is hard for image-based optical flow algorithms to compute motion vectors in bad lighting conditions, obtaining more distortions in the output flow map. 
On the other hand, lidarFlow in the final row provides improved optical flow in such challenging conditions where there are less distortions than rgbFlow at night, and no shadow-based distortion because LiDAR does not capture color textures. 
Yet, due to the sparsity of the LiDAR point clouds which increases with further objects, motion of far objects is  modelled with difficulty compared to flow maps from dense RGB images.

\subsection{Network Architecture}

In this section, we detail our baseline architectures, and the different implemented fusion approaches. \\

%\vspace{1mm}
\textbf{Baseline Architecture:} 
We set our baseline based on \cite{gamal2018shuffleseg}, which presents an encoder-decoder schema. Our encoder is responsible of extracting features before the upsampling phase done by the decoder and is based on \cite{zhang2018shufflenet}, which uses point-wise group convolutions and channel shuffling. This in turn reduces computation cost at a high accuracy level which is perfect for a real-time application such as needed on autonomous driving systems. Our decoder is based on \cite{long2015fully} which is composed of three deconvolution layers that provide the final output image size.  This approach has the advantage of low complexity as well as provides a lightweight network architecture able to fit on autonomous driving embedded platforms. Detailed analysis of efficient design techniques for segmentation is discussed in \cite{siam2018rtseg, briot2018analysis}.
Two classes are used to train the network, i.e, Moving and Non-Moving. In addition to the static objects, background pixels are considered as Non-Moving, therefore the number of static pixels exceeds the number of moving pixels. Weighted cross-entropy is used to overcome this class imbalance problem. We make use of this architecture to evaluate a baseline performance using RGB images only. 

%\vspace{1mm}
\textbf{Early Fusion:} 
Early-Fusion is referred to as data-fusion where fusion is done on the data level before any feature extraction. The same baseline network architecture is utilized in this case, however the input data is concatenated at the very beginning. 
%however the input data is concatenated where the corresponding tensor weights are initialized randomly.
This architecture has the advantage of low-complexity compared to Mid-Fusion approach, as the number of weights is kept similar to the baseline architecture being the main difference on the input layer only. 

%\vspace{1mm}
\textbf{Mid Fusion:} 
Mid-Fusion refers to feature-level-fusion where features are extracted from each input separately using an encoder that is exclusive to each input. Fusion is done by concatenating feature maps that are generated from each stream before upsampling in the decoder. This architecture provides the best fusion performance, however it has higher cost than early-fusion as the number of weights in the encoder part is doubled. 

\begin{figure*}[tb]
    \centering
    \includegraphics[width=.9\textwidth]{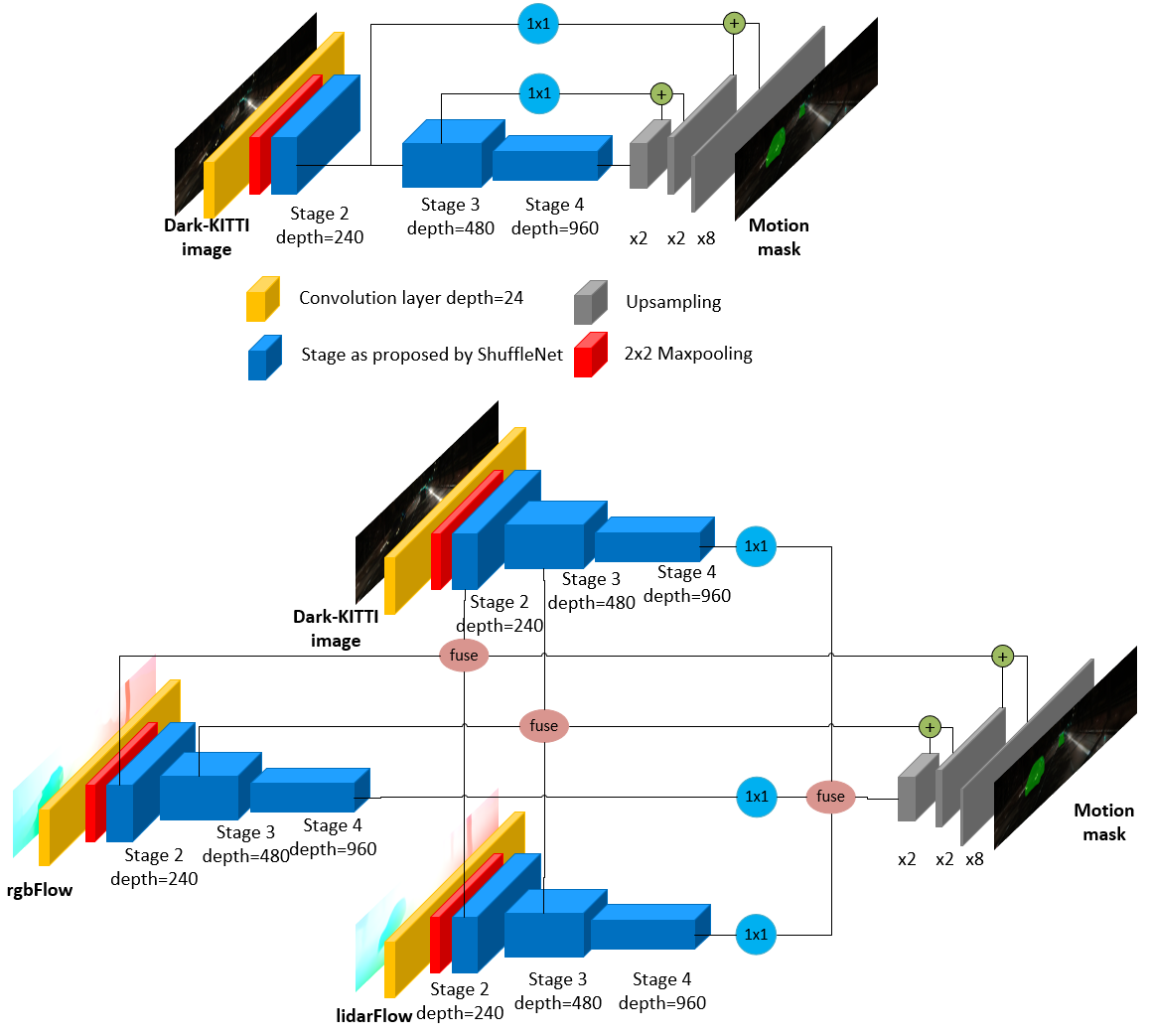}
    \caption{\textbf{Top:} Baseline architecture based on \cite{gamal2018shuffleseg}. \textbf{Bottom:} Proposed fusion architecture.
    \label{fig:threeStream}}
    \vspace{-0.4cm}
\end{figure*}

%\vspace{1mm}
\textbf{Hybrid Fusion:}
This architecture makes use of both early and mid-Fusion. We use it in various experiments as illustrated in Table \ref{tab:accuracy}, in an attempt to maximize the benefit of the input modalities while avoiding too much complexity for the model at hand. For instance, we fuse 4 inputs, i.e, RGB, rgbFlow, lidarFlow, LiDAR depth  through early-fusion in one branch between RGB and rgbFlow, and early-fusion in another branch for LiDAR depth and lidarFlow. The output of both branches is fused through Mid-Fusion. 

%\vspace{1mm}
\textbf{Proposed Architecture:}
We aim at finding the best schema to combine RGB images, rgbFlow and lidarFlow. For that purpose, we construct a three-stream mid-Fusion network which has three encoders for RGB, rgbFlow and lidarFlow separately. We evaluate this approach on KITTI and Dark-KITTI datasets, where results demonstrate the improved performance on both datasets as detailed in section \ref{sec:results}.

\section{Experiments} \label{sec:exps}
% \begin{figure*}[tb]
%     \centering
%     % \includegraphics[width=0.99\linewidth]{images/SAW_definition}
%     % \includegraphics[width=0.89\linewidth]{images/SAW_definition}
%     \includegraphics[width=0.95\linewidth]{images/SAW_definition}
%     \caption{From left to right: a) soiled camera lens mounted to the car body; b) the image quality of the soiled camera from the previous image; c) an example of image soiled by a heavy rain.}
%     \label{fig:saw_definition}
% \end{figure*}

\subsection{Experimental Setup} \label{setup}
In all our experiments, ShuffleSeg \cite{gamal2018shuffleseg} model was used with pretrained ShuffleNet encoder on Cityscapes dataset for Semantic Segmentation. For the decoder part,  FCN8s decoder has been utilized with randomly initialized weights. L2
regularization with weight decay rate of $5e^{-4}$ and Batch Normalization are incorporated. We trained all our models End-To-End with weighted binary cross-entropy loss for 200 epochs and batch size 6. Adam optimizer is used with learning rate of $1e^{-4}$.
For inputs with number \textit{n} of channels lower than 3, we discarded the difference of depth from the filters of the first convolutional layer. For the rest of inputs, we increased the depth of the filters by the first \textit{n} channels of the single filter to match the first layer with the new input shape, initializing the corresponding weights randomly.

\subsection{Experimental Results} \label{sec:results}
\begin{figure*}[t!]
\captionsetup[subfigure]{labelformat=empty}
\centering
\begin{subfigure}{.498\textwidth}
    \includegraphics[width=\textwidth]{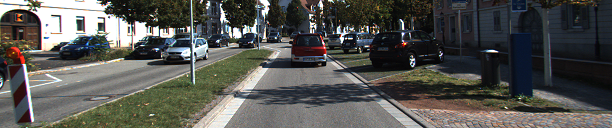}
    \vspace{-1cm}
    \caption{\textcolor{white}{(a)}}
 %   \caption{\textcolor{black}{KITTI input image}\newline}
\end{subfigure}%
\hfill
\begin{subfigure}{.498\textwidth}
    \includegraphics[width=\textwidth]{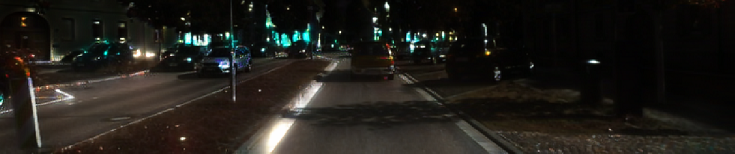}
    \vspace{-1cm}
    \caption{\textcolor{white}{(b)}}
 %   \caption{\textcolor{black}{Dark-KITTI input image}\newline}
\end{subfigure}%
\quad
\begin{subfigure}{.498\textwidth}
    \includegraphics[width=\textwidth]{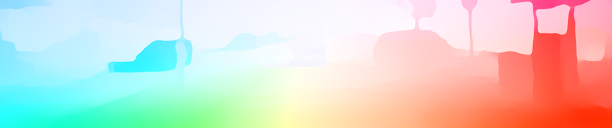}
    \vspace{-1cm}
    \caption{\textcolor{black}{(c)}}
 %   \caption{\textcolor{black}{Day Input rgbFlow} \newline}
\end{subfigure}%
\hfill
\begin{subfigure}{.498\textwidth}
    \includegraphics[width=\textwidth]{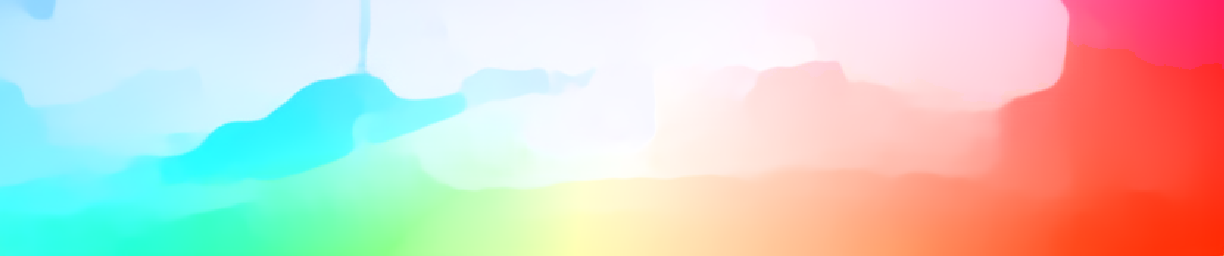}
    \vspace{-1cm}
    \caption{\textcolor{black}{(d)}}
 %   \caption{\textcolor{black}{Night Input rgbFlow} \newline}
\end{subfigure}%
\quad
\begin{subfigure}{.498\textwidth}
    \includegraphics[width=\textwidth]{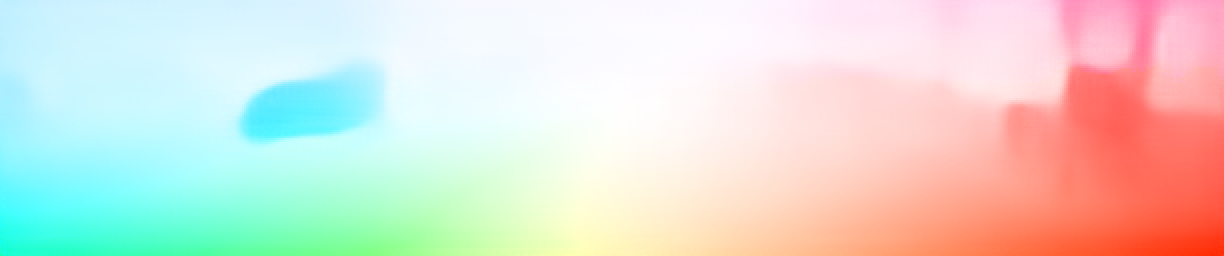}
    \vspace{-1cm}
    \caption{\textcolor{black}{(e)}}
%    \caption{\textcolor{black}{Day Input rgbFlow} \newline}
\end{subfigure}%
\hfill
\begin{subfigure}{.498\textwidth}
    \includegraphics[width=\textwidth]{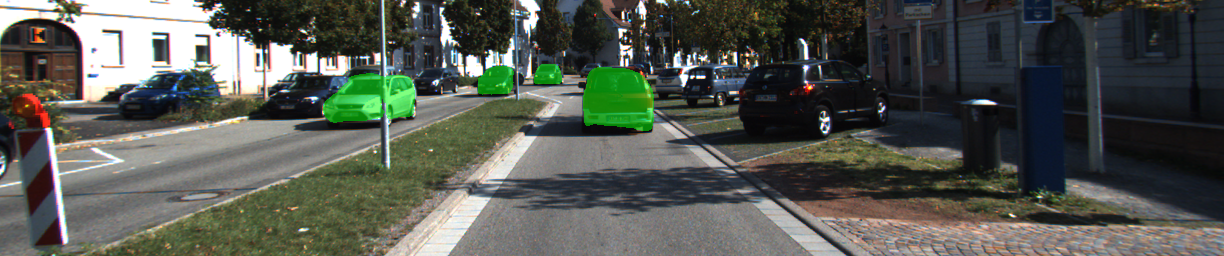}
    \vspace{-1cm}
    \caption{\textcolor{white}{(f)}}
%    \caption{\textcolor{black}{Night Input rgbFlow} \newline}
\end{subfigure}%
\quad
\begin{subfigure}{.498\textwidth}
    \includegraphics[width=\textwidth]{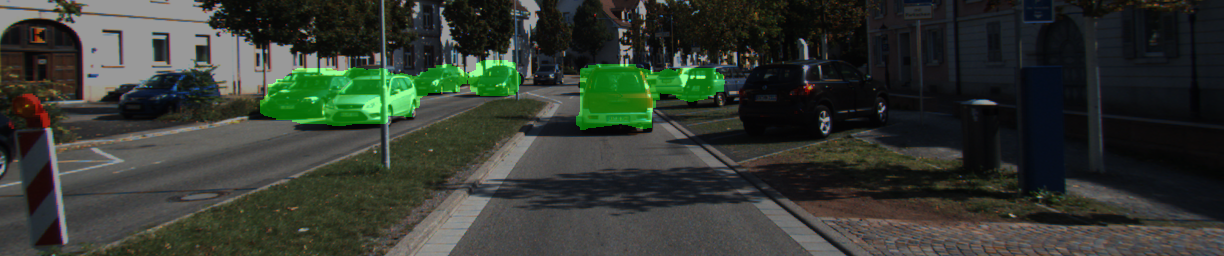}
    \vspace{-1cm}
    \caption{\textcolor{white}{(g)}}
%    \caption{\textcolor{black}{RGB only} \newline}
\end{subfigure}%
\hfill
\begin{subfigure}{.498\textwidth}
    \includegraphics[width=\textwidth]{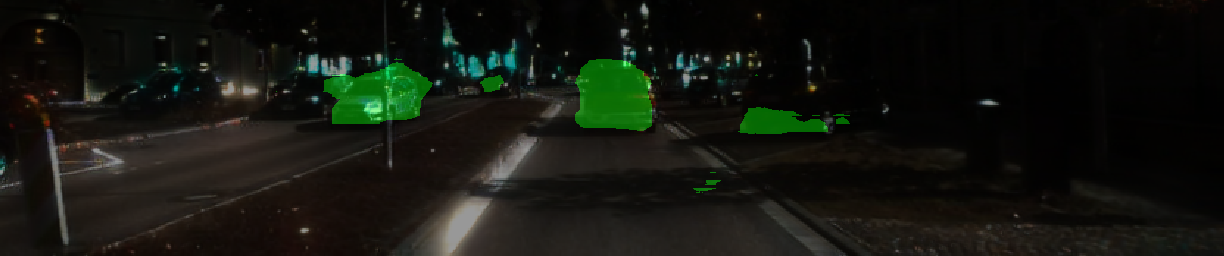}
    \vspace{-1cm}
    \caption{\textcolor{white}{(h)}}
%    \caption{\textcolor{black}{RGB only}\newline}
\end{subfigure}
\quad
\begin{subfigure}{.498\textwidth}
    \includegraphics[width=\textwidth]{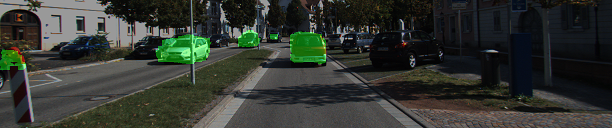}
    \vspace{-1cm}
    \caption{\textcolor{white}{(i)}}
%    \caption{\textcolor{black}{RGB + rgbFlow\_Day}\newline}
\end{subfigure}%
\hfill
\begin{subfigure}{.498\textwidth}
    \includegraphics[width=\textwidth]{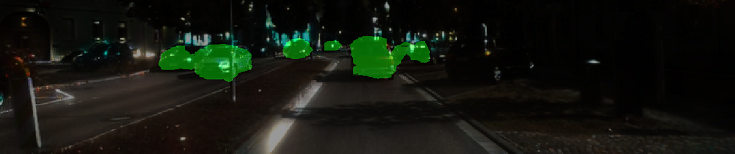}
    \vspace{-1cm}
    \caption{\textcolor{white}{(j)}}
%    \caption{\textcolor{black}{RGB + rgbFlow\_Night}\newline}
\end{subfigure}
\quad
\begin{subfigure}{.498\textwidth}
    \includegraphics[width=\textwidth]{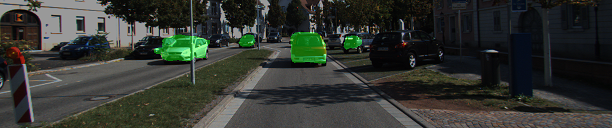}
    \vspace{-1cm}
    \caption{\textcolor{white}{(k)}}
 %   \caption{\textcolor{black}{RGB + lidarFlow}\newline}
\end{subfigure}%
\hfill
\begin{subfigure}{.498\textwidth}
    \includegraphics[width=\textwidth]{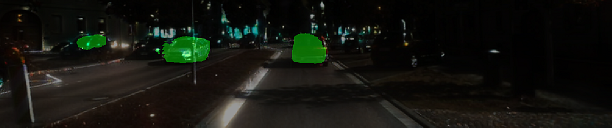}
    \vspace{-1cm}
    \caption{\textcolor{white}{(l)}}
 %   \caption{\textcolor{black}{RGB + lidarFlow}\newline}
\end{subfigure}%
\quad
\begin{subfigure}{.498\textwidth}
    \includegraphics[width=\textwidth]{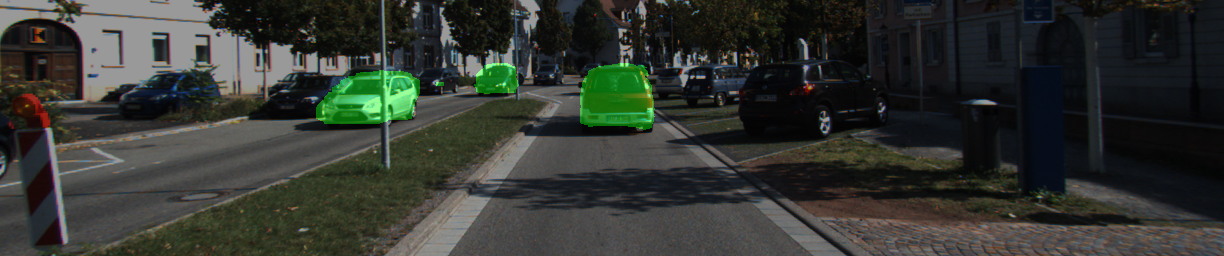}
    \vspace{-1cm}
    \caption{\textcolor{white}{(m)}}
 %   \caption{\textcolor{black}{RGB + rgbFlow\_Day + lidarFlow}}
\end{subfigure}%
\hfill
\begin{subfigure}{.498\textwidth}
    \includegraphics[width=\textwidth]{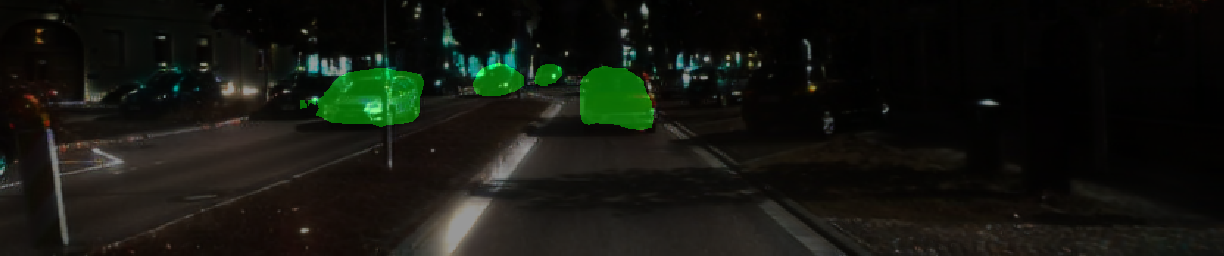}
    \vspace{-1cm}
    \caption{\textcolor{white}{(n)}}
 %   \caption{\textcolor{black}{RGB + rgbFlow\_Night + lidarFlow} }
\end{subfigure}%
\quad

    \caption{Qualitative comparison of our algorithm on KITTI and Dark-KITTI datasets. First column shows inputs and results on KITTI while second shows results on Dark-KITTI. \textbf{(a),(b)} show the input RGB images. \textbf{(c),(d)} show rgbFlow. \textbf{(e)} shows lidarFlow. \textbf{(f)} shows Ground Truth. \textbf{(g),(h)} show output using RGB-only. \textbf{(i),(j)} show output of (RGB + rgbFlow). \textbf{(k),(l)} show output of (RGB + lidarFlow). \textbf{(m),(n)} show output of (RGB + rgbFlow + lidarFlow).}
    \label{fig:qualitativeEval}
    \vspace{-0.4cm}
\end{figure*}

We provide a table of quantitative results for both day and night images evaluated on KITTI and Dark-KITTI datasets. Qualitative evaluation on both datasets is illustrated in Figure \ref{fig:qualitativeEval}. 

Table \ref{tab:accuracy} demonstrates our results using mean IoU metric for both moving and background class and IoU for the moving objects, in addition to class-wise IoU for ``Moving" class. We refer to early-fusion by ``x" while ``+" denotes mid-fusion where both of them together imply hybrid fusion. RGB-only experiments serve as a baseline for comparative purpose where we evaluate our network architecture to segment moving objects using color information only without either explicit or implicit motion signal for the network. Significant improvement for 13\% in moving class IoU has been observed after fusion with optical flow, which is consistent with previous conclusions in \cite{siam2017modnet,siam2018real}. We attempt to minimize complexity through early-fusion architecture as we focus on real-time architecture for autonomous driving. However it is found that early-fusion architecture only (RGB x rgbFlow) is not capable of extracting the required features compared to Mid-Fusion which is consistent with other literature such as \cite{rashed2019motion,dewan2016motion}. Thus we continue our experiments using Mid or Hybrid fusion. Mid-Fusion experiment with rgbFlow (RGB + rgbFlow) serves as a comparison baseline as well because our motivation is to evaluate the augmentation of motion information from LiDAR sensor. (RGB + lidarFlow) shows improved performance over RGB-only, however overall accuracy is still below (RGB + rgbFlow). 

Nevertheless, we argue that both lidarFlow and rgbFlow are complementary to each other where rgbFlow benefits from dense color information which is helpful to understand motion for far objects, however illumination plays a great role in the quality of optical flow from RGB images. On the other hand, lidarFlow might not provide the best motion estimate of far objects due to increased sparsity when the objects are far away, however, it is illumination independent due to relying on TOF concept which is perfect for low illumination scenes motion estimation. Our approach is proven experimentally through the (RGB + rgbFlow + lidarFlow) experiment where we obtain absolute improvement of 4\% and relative improvement of 10\% in IoU over (RGB + rgbFlow). We attempt to fuse optical flow information before feature extraction through hybrid-fusion (RGB + (rgbFlow x lidarFlow)), in addition to experimentation of leveraging depth points through a two stream approach (RGB x rgbFlow) + (LiDAR x lidarFlow). LidarFlow augmentation shows improvement in results over the baseline (RGB + rgbFlow) which proves our approach. However, our three-stream approach gives the network more flexibility to combine features from each input for maximum accuracy. 

Implicit motion learning has been explored in (RGB time \textit{t} x RGB time \textit{t+1}) + (LiDAR depth \textit{t} x LiDAR depth \textit{t+1}) where the network is expected to learn motion implicitly without optical flow computation. An improvement is observed compared to RGB-only baseline however we obtain degradation in performance compared to explicit motion learning, and this is expected because the network learns to model motion vectors implicitly in addition to its original task which is MOD. We evaluate our approach on KITTI dataset, and we show that lidarFlow augmentation improves accuracy of moving objects even in high-illumination images where 2\% improvement in IoU is observed compared to camera-only solution. These results demonstrate that our approach is beneficial for motion segmentation task regardless of illumination parameter which was a drawback in the previous literature. 

\begin{table}[t]
\caption{Quantitative results on KITTI and Dark-KITTI. ``+" refers to Mid-Fusion. ``x" refers to Early-Fusion. Both together refer to Hybrid-Fusion.}
\label{tab:accuracy}
\begin{adjustbox}{width=.48\textwidth}
\begin{tabular}{|l|l|l|}
\hline
\multicolumn{1}{|c|}{\textbf{Type}} & \multicolumn{1}{c|}{\textbf{mIoU}} & \multicolumn{1}{c|}{\textbf{Moving IoU}} \\ \hline
\multicolumn{3}{|c|}{Dark-KITTI} \\ \hline

RGB-only  & 62.6 & 26.5  \\ \hline
RGB + rgbFlow  & 69.2 & 39.5  \\ \hline
RGB x rgbFlow  & 61.68 & 24.86  \\ \hline
RGB + lidarFlow & 68.7 & 38.5  \\ \hline
% RGB x rgbFlow-mag x lidarFlow-mag & 67.15 & 35.38  \\\hline

\begin{tabular}[c]{@{}l@{}}(RGB time \textit{t} x RGB time \textit{t+1}) +  \\  (LiDAR depth \textit{t} x LiDAR depth \textit{t+1})\end{tabular}
 & 66.26 & 33.83 \\\hline
 
 \begin{tabular}[c]{@{}l@{}}(RGB x rgbFlow) + (LiDAR depth x  \\   lidarFlow)\end{tabular}
 
 & 69.92 & 40.93 \\\hline
RGB + (rgbFlow x lidarFlow) & 69.8 & 40.75 \\\hline
\textbf{RGB + rgbFlow + lidarFlow} & \textbf{71.2} & \textbf{43.5}  \\\hline

\multicolumn{3}{|c|}{KITTI}                                                                                              \\ \hline
RGB-only  & 65.6 & 32.7  \\ \hline
RGB + rgbFlow  & 74.24 & 49.36 \\ \hline
RGB + lidarFlow & 70.27 & 41.64  \\ \hline

\begin{tabular}[c]{@{}l@{}}(RGB time \textit{t} x RGB time \textit{t+1}) +  \\  (LiDAR depth \textit{t} x LiDAR depth \textit{t+1})\end{tabular}
 & 66.68 & 34.67 \\\hline
RGB + (rgbFlow x lidarFlow) & 72.21 & 45.45 \\\hline
\textbf{RGB + rgbFlow + lidarFlow} & \textbf{75.3} & \textbf{51.46}  \\\hline
\end{tabular}
\vspace{-0.4cm}
\end{adjustbox}
\end{table}

\begin{table}[tb]
\caption{Comparison between the tested architectures for MOD task. Frame per second (fps) is used as a metric to evaluate real-time performance. Evaluation is performed on 256x1224 resolution images on Titan X Pascal GPU.}
\label{tab:realTimePerf}
\begin{adjustbox}{width=.48\textwidth}
\begin{tabular}{|l|l|l|}
\hline
\multicolumn{1}{|c|}{\textbf{Type}} & \multicolumn{1}{c|}{\textbf{fps}} \\ \hline
Baseline architecture  & 40  \\ \hline
Two-Stream Mid-Fusion architecture  & 25  \\ \hline
Three-Stream proposed Mid-Fusion architecture  & 18  \\ \hline
\end{tabular}
\end{adjustbox}
\end{table}

% \begin{figure*}[t!]
% \captionsetup[subfigure]{labelformat=empty}
% \centering

%     \includegraphics[width=.49\textwidth]{images/samples/inputs/5.png}
%     \vspace{-1cm}
%     \caption{\textcolor{white}{(a)}}
%     \includegraphics[width=.49\textwidth]{images/samples/inputs/6.png}

% % \begin{subfigure}{\textwidth}
% %     \includegraphics[width=.48\textwidth]{images/samples/inputs/5.png}
% % \end{subfigure}%
% % \quad
% % \begin{subfigure}{\textwidth}
% %     \includegraphics[width=.48\textwidth]{images/samples/inputs/6.png}
% % \end{subfigure}%

% \end{figure*}

Figure \ref{fig:qualitativeEval} demonstrates our results obtained in Table \ref{tab:accuracy}. The first column shows results of our algorithm on KITTI dataset and the second one reports Dark-KITTI results. The input RGB images are shown in the first row. The second row shows the input optical flow maps of KITTI and Dark-KITTI. The third row shows lidarFlow map and ground truth. Fourth row reports results of MOD using only color information as an input. It is shown that the network only learned to segment the cars and not the moving cars as shown in both KITTI and Dark-KITTI results. Some of the parked cars are not segmented because it might be implicitly learned that cars in that position are not interesting. However, this is not based on motion information, and this is expected because there is no motion information given to the algorithm either explicitly or implicitly. 

In Dark-KITTI, only two vehicles are segmented because of low illumination where it is even hard to segment them using human eyes. Fusion with optical flow in the fifth row has improved results significantly on both datasets however, there are too many false positives in Dark-KITTI dataset as in (h) due to inaccurate optical flow because of low illumination of the scene. The sixth row shows results of fusion of color information from camera and motion information from LiDAR. Results show improved performance over (RGB + rgbFlow) especially on Dark-KITTI dataset. This is due to illumination independent optical flow from lidarFlow \cite{vaquero2018hallucinating}. However, far objects are still not captured correctly due to increased sparsity with far objects. The seventh row demonstrate the results of our proposed architecture which combines color information, motion information from both camera and LiDAR sensors. Results show the benefit of fusion where the network was able to maximize accuracy from both sensors and segment the scene moving objects. 

\begin{figure*}[ht!]
\captionsetup[subfigure]{labelformat=empty}
\centering
\begin{subfigure}{.498\textwidth}
    \includegraphics[width=\textwidth]{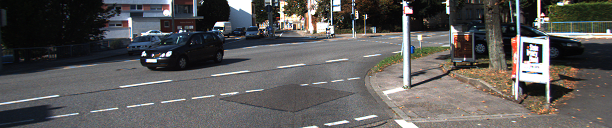}
    \vspace{-1cm}
    \caption{\textcolor{white}{(a)}}
 %   \caption{\textcolor{black}{KITTI input image}\newline}
\end{subfigure}%
% \begin{picture}(1,0.55038404)%
%   \put(0,0){\includegraphics[width=.4\textwidth]{17D2J}}%
%   \put(150,50){x axis}%
% \end{picture}%
\hfill
\begin{subfigure}{.498\textwidth}
    \includegraphics[width=\textwidth]{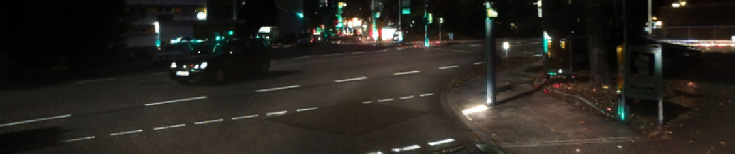}
    \vspace{-1cm}
    \caption{\textcolor{white}{(b)}}
 %   \caption{\textcolor{black}{Dark-KITTI input image}\newline}
\end{subfigure}%
\quad
\begin{subfigure}{.498\textwidth}
    \includegraphics[width=\textwidth]{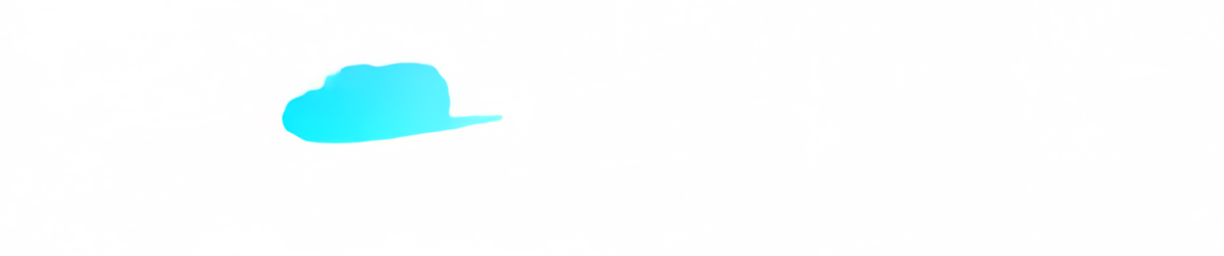}
    \vspace{-1cm}
    \caption{\textcolor{black}{(c)}}
 %   \caption{\textcolor{black}{Input rgbFlow\_Day}\newline}
\end{subfigure}%
\hfill
\begin{subfigure}{.498\textwidth}
    \includegraphics[width=\textwidth]{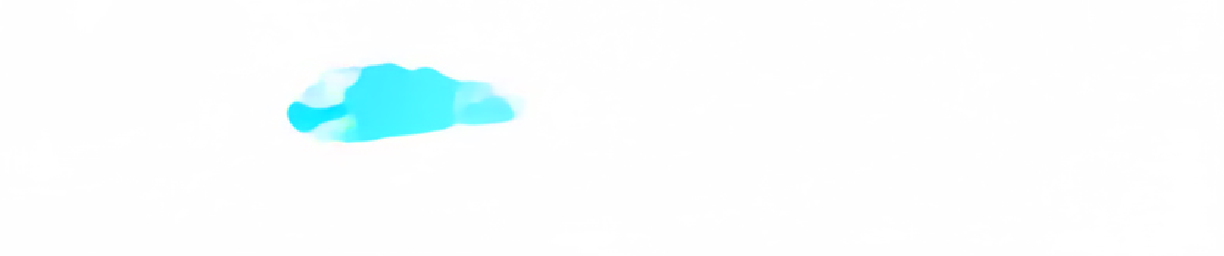}
    \vspace{-1cm}
    \caption{\textcolor{black}{(d)}}
 %   \caption{\textcolor{black}{Input rgbFlow\_Night}\newline}
\end{subfigure}%
\quad
\begin{subfigure}{.498\textwidth}
    \includegraphics[width=\textwidth]{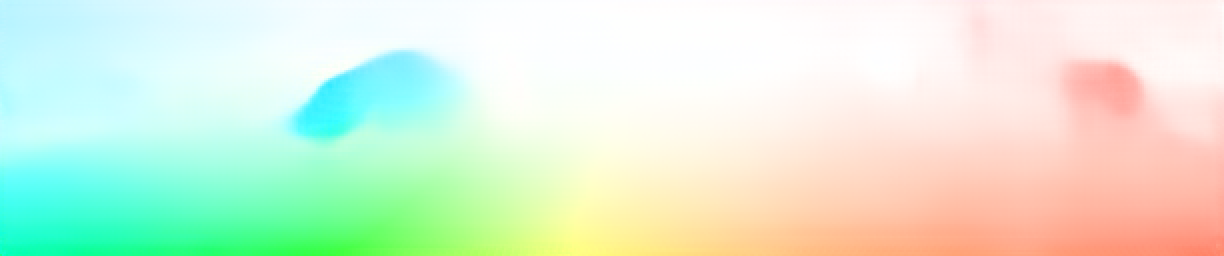}
    \vspace{-1cm}
    \caption{\textcolor{black}{(e)}}
 %   \caption{\textcolor{black}{Input LiDARFlow}\newline}
\end{subfigure}%
\hfill
\begin{subfigure}{.498\textwidth}
    \includegraphics[width=\textwidth]{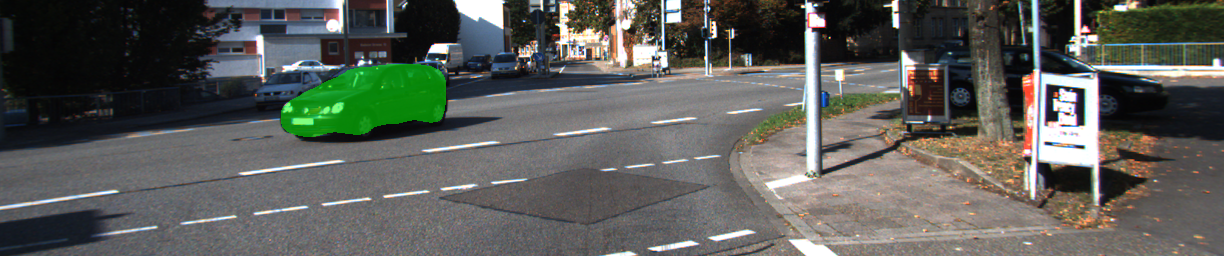}
    \vspace{-1cm}
    \caption{\textcolor{white}{(f)}}
 %   \caption{\textcolor{black}{Ground-Truth}\newline}
\end{subfigure}%
\quad
\begin{subfigure}{.498\textwidth}
    \includegraphics[width=\textwidth]{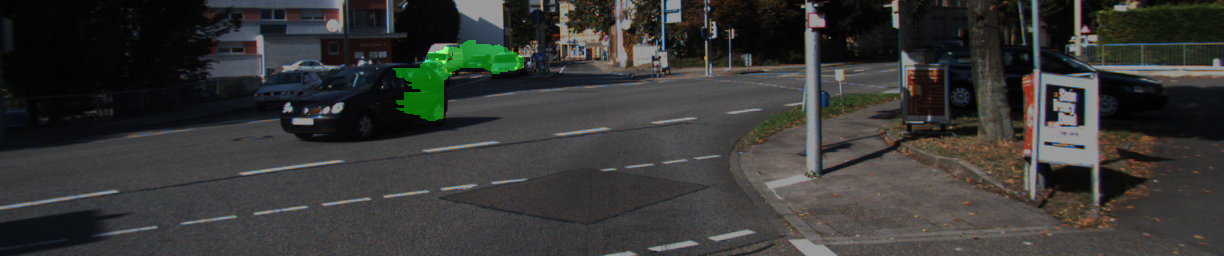}
    \vspace{-1cm}
    \caption{\textcolor{white}{(g)}}
 %   \caption{\textcolor{black}{RGB only} \newline}
\end{subfigure}%
\hfill
\begin{subfigure}{.498\textwidth}
    \includegraphics[width=\textwidth]{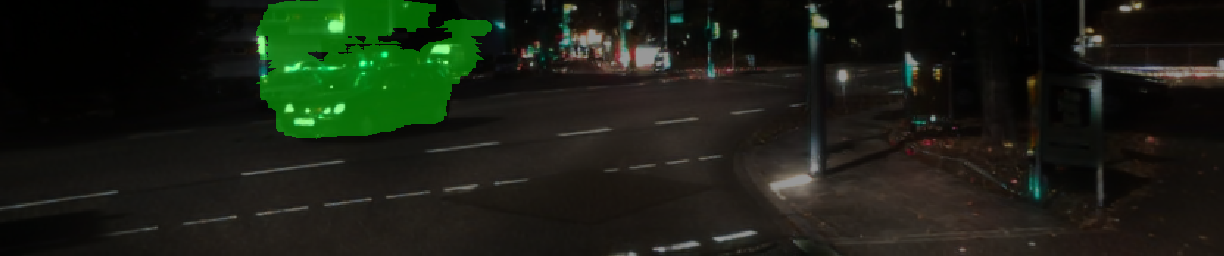}
    \vspace{-1cm}
    \caption{\textcolor{white}{(h)}}
 %   \caption{\textcolor{black}{RGB only}\newline}
\end{subfigure}
\quad
\begin{subfigure}{.498\textwidth}
    \includegraphics[width=\textwidth]{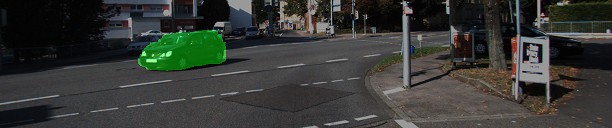}
    \vspace{-1cm}
    \caption{\textcolor{white}{(i)}}
 %   \caption{\textcolor{black}{RGB + rgbFlow\_Day}\newline}
\end{subfigure}%
\hfill
\begin{subfigure}{.498\textwidth}
    \includegraphics[width=\textwidth]{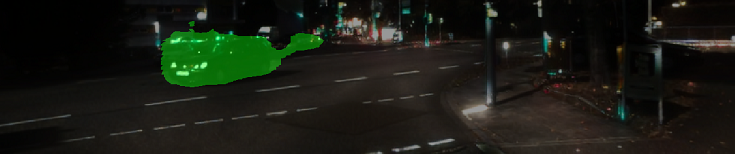}
    \vspace{-1cm}
    \caption{\textcolor{white}{(j)}}
 %   \caption{\textcolor{black}{RGB + rgbFlow\_Night}\newline}
\end{subfigure}
\quad
\begin{subfigure}{.498\textwidth}
    \includegraphics[width=\textwidth]{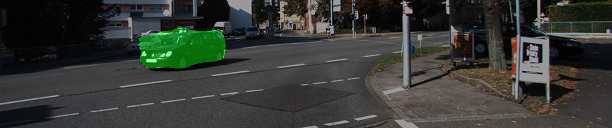}
    \vspace{-1cm}
    \caption{\textcolor{white}{(k)}}
 %   \caption{\textcolor{black}{RGB + lidarFlow}\newline}
\end{subfigure}%
\hfill
\begin{subfigure}{.498\textwidth}
    \includegraphics[width=\textwidth]{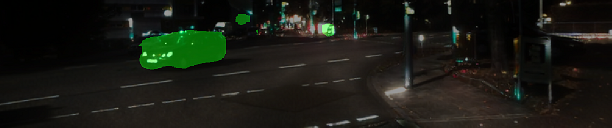}
    \vspace{-1cm}
    \caption{\textcolor{white}{(l)}}
 %   \caption{\textcolor{black}{RGB + lidarFlow}\newline}
\end{subfigure}%
\quad
\begin{subfigure}{.498\textwidth}
    \includegraphics[width=\textwidth]{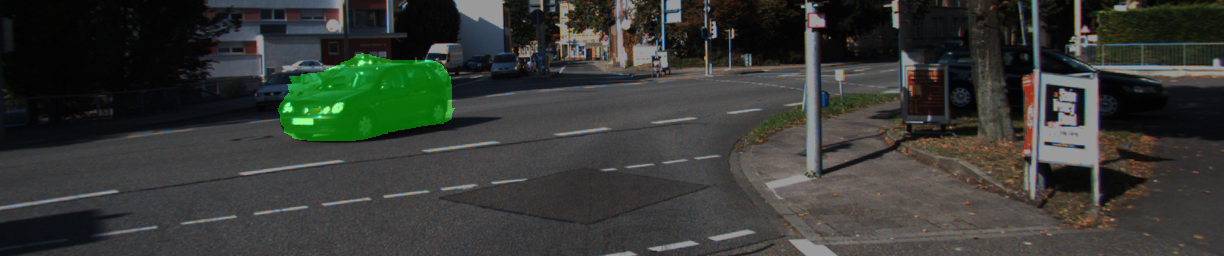}
    \vspace{-1cm}
    \caption{\textcolor{white}{(m)}}
 %   \caption{\textcolor{black}{RGB + rgbFlow\_Day + lidarFlow}}
\end{subfigure}%
\hfill
\begin{subfigure}{.498\textwidth}
    \includegraphics[width=\textwidth]{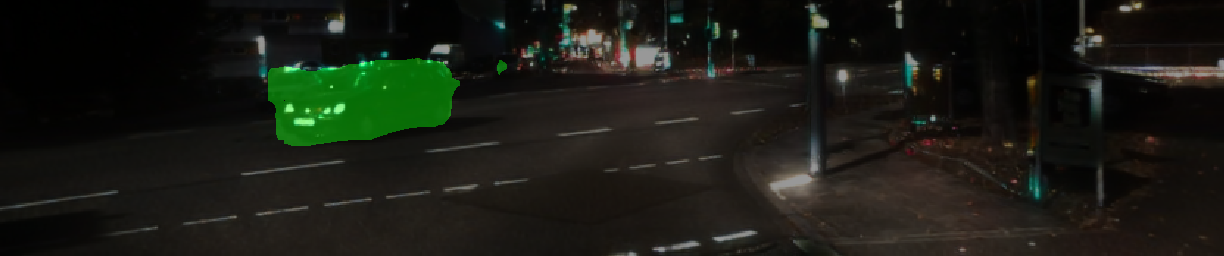}
    \vspace{-1cm}
    \caption{\textcolor{white}{(n)}}
 %   \caption{\textcolor{black}{RGB + rgbFlow\_Night + lidarFlow} }
\end{subfigure}%
\quad

    \caption{Qualitative comparison of our algorithm on KITTI and Dark-KITTI datasets. First column shows inputs and results on KITTI while second shows results on Dark-KITTI. \textbf{(a),(b)} show the input RGB images. \textbf{(c),(d)} show rgbFlow. \textbf{(e)} shows lidarFlow. \textbf{(f)} shows Ground Truth. \textbf{(g),(h)} show output using RGB-only. \textbf{(i),(j)} show output of (RGB + rgbFlow). \textbf{(k),(l)} show output of (RGB + lidarFlow). \textbf{(m),(n)} show output of (RGB + rgbFlow + lidarFlow).}
    \label{fig:qualitativeEval_2}
    \vspace{-0.4cm}
\end{figure*}

Figure \ref{fig:qualitativeEval_2} shows an example of failure of our algorithm where it is shown that output without augmentation of lidarFlow in a high-illumination image is slightly better than using lidarFlow. In this sample, the ego-vehicle is static, and there is only one car that is moving in the scene as illustrated in ground truth. The rgbFlow obtained during day which is shown in (c) provides maximum accuracy when it is fused with RGB as illustrated in (i). Due to inaccurate motion map obtained from LiDAR which is shown in (e), some distortions took place when this input was fused with rgbFlow. This is illustrated in (m) compared to (i). However, the distortion is minimal where the network is still able to learn motion mask correctly even with noisy lidarFlow. Moreover, overall moving IoU has improved with 2\% after augmentation of lidarFlow with rgbFlow for high-illumination images as illustrated in Table \ref{tab:accuracy}. On the other hand, for Dark-KITTI dataset, the fusion with the noisy lidarFlow improves performance of low-illumination images as illustrated in (n) compared to (j) which provides that our algorithm is illumination independent and works perfectly in all lighting conditions.
Table \ref{tab:realTimePerf} shows real-time evaluation performance of our algorithm. Our proposed model runs 18 fps which is suitable for real-time application such as the autonomous driving. The results are reported using images of resolution 256x1224 on Titan X Pascal GPU.

\section{Conclusions} \label{sec:conclusions}
%\vspace{-0.2cm}
We explored the impact of leveraging LiDAR sensor for understanding scene motion for MOD especially for low-illumination autonomous driving conditions. We created our own dataset Dark-KITTI to evaluate our algorithm in low-light conditions by extending the public MOD dataset \cite{siam2017modnet}.  We constructed different fusion algorithms to empirically study best fusion methodology. We proposed a novel architecture that fuses color signal with motion information that is captured from both camera and LiDAR sensors. Our model is evaluated on both night and day images and we obtain improved performance in both of them. The proposed architecture is designed for real-time performance for autonomous driving application where our most complex algorithm runs at 18 fps. We hope that this study encourages further research in construction of better fusion networks.

\clearpage % Remove me if the layout gets fucked up!

% %%%%%%%%%%%%%%%%%%%%%%%%%%%%%%%%%%%%
% References
% %%%%%%%%%%%%%%%%%%%%%%%%%%%%%%%%%%%%
{\small
\bibliographystyle{ieee}
\bibliography{references/egbib}
}

\end{document}